\documentclass[nohyperref]{article}

\usepackage{microtype}
\usepackage{graphicx}
\usepackage{subfigure}
\usepackage{booktabs} 
\usepackage{ulem}

\usepackage{longtable}
\usepackage{natbib}
\defcitealias{sandbox2}{GitHub Repository}
\defcitealias{webdriver}{ChromeDriver Security Considerations}

\usepackage{ltcaption}

\usepackage{hyperref}

\usepackage[accepted]{icml2022}

\usepackage{amsmath}
\usepackage{amssymb}
\usepackage{mathtools}
\usepackage{amsthm}

\usepackage[capitalize,noabbrev]{cleveref}

\theoremstyle{plain}

\theoremstyle{definition}

\theoremstyle{remark}

\usepackage[textsize=tiny]{todonotes}

\icmltitlerunning{A Data-Driven Approach for Learning to Control Computers}

\begin{document}

\twocolumn[
\icmltitle{A Data-Driven Approach for Learning to Control Computers}

\icmlsetsymbol{equal}{*}

\begin{icmlauthorlist}
\icmlauthor{Peter Humphreys}{dm}
\icmlauthor{David Raposo}{dm}
\icmlauthor{Toby Pohlen}{dm}
\icmlauthor{Gregory Thornton}{dm}
\icmlauthor{Rachita Chhaparia}{dm}
\icmlauthor{Alistair Muldal}{dm}
\icmlauthor{Josh Abramson}{dm}
\icmlauthor{Petko Georgiev}{dm}
\icmlauthor{Adam Santoro}{dm}
\icmlauthor{Timothy Lillicrap}{dm}
\end{icmlauthorlist}

\icmlaffiliation{dm}{DeepMind, London, United Kingdom}

\icmlcorrespondingauthor{Peter Humphreys \textless{}peterhumphreys@deepmind.com\textgreater{}, Timothy Lillicrap}{countzero@deepmind.com}

\icmlkeywords{Machine Learning, ICML}

\vskip 0.3in
]



\printAffiliationsAndNotice{}

\begin{abstract}
It would be useful for machines to use computers as humans do so that they can aid us in everyday tasks. This is a setting in which there is also the potential to leverage large-scale expert demonstrations and human judgements of interactive behaviour, which are two ingredients that have driven much recent success in AI. Here we investigate the setting of computer control using keyboard and mouse, with goals specified via natural language. Instead of focusing on hand-designed curricula and specialized action spaces, we focus on developing a scalable method centered on reinforcement learning combined with behavioural priors informed by actual human-computer interactions. We achieve state-of-the-art and human-level mean performance across all tasks within the MiniWob\texttt{++} benchmark, a challenging suite of computer control problems, and find strong evidence of cross-task transfer. These results demonstrate the usefulness of a unified human-agent interface when training machines to use computers. Altogether our results suggest a formula for achieving competency beyond MiniWob\texttt{++} and towards controlling computers, in general, as a human would. 
\end{abstract}

\section{Introduction}

Recent work on natural language \citep{brown2020language,rae2021scaling}, code production \citep{chen2021evaluating}, and multimodal interactive behaviour in 3D simulated worlds \citep{team2021creating} has produced models with remarkable expressivity, context awareness, and general knowledge. This research is a striking demonstration of the power of two ingredients: a rich, compositional output space that is congruent between machines and humans, and large amounts of human data and judgements to inform machine behaviour. 

One domain that has these two ingredients but has received relatively less attention is \emph{digital device control} \citep{shi2017world,liu2018reinforcement,nakano2021webgpt,chen2021evaluating, shvo2021appbuddy, li2020mapping, toyama2021androidenv, gur2021environment}, which comprises the use of digital devices to accomplish a myriad of inherently useful tasks. Because of its near exclusive use of digital information, this domain scales well in regards to data acquisition and parallelization of control (compared to, say, robotics or fusion reactors). It also combines diverse, multimodal inputs with expressive, composable, and human-compatible affordances. In this work we focus on \textit{computer control} using keyboard and mouse, with pixel and Document Object Model (DOM) observations.

\begin{figure}[h!]
    \centering
    \includegraphics[width=0.46\textwidth]{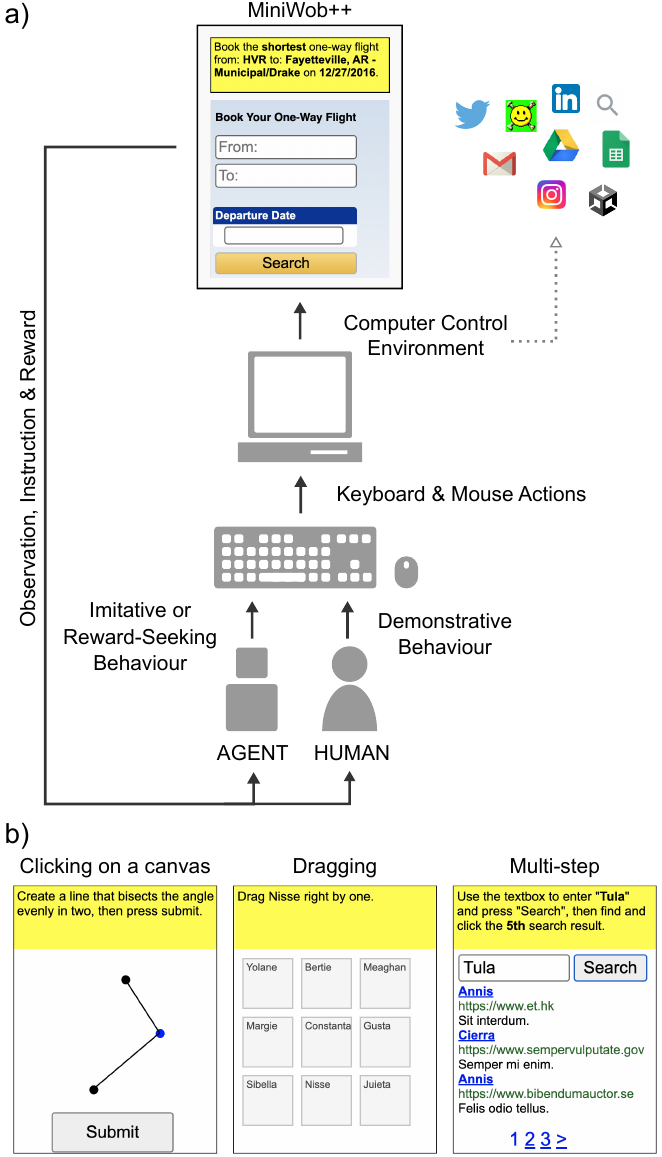}
    \vspace{-0.0cm}
    \caption{\textbf{Computer Control Environment running MiniWob\texttt{++}}. Both humans and agents control computers using a keyboard and mouse, with humans providing demonstrative behaviour used for behavioural cloning, and agents trained to mimic this behaviour or to behave in pursuit of reward. Humans and agents attempt to solve the MiniWob\texttt{++} task suite, which comprises instruction-following tasks requiring clicking, typing, dragging, form-filling, and other such basic interactions (depicted is the ``book-flight'' task). b) Examples of MiniWob tasks requiring different forms of interaction.}
    \label{fig:concept}
\end{figure}

A useful benchmark for initial investigations of computer control is the MiniWob\texttt{++} task suite \citep{shi2017world,liu2018reinforcement}, which comprises a set of instruction following tasks that require clicking, typing, form-filling, and other such basic computer interactions (Fig.~\ref{fig:concept}b). MiniWob\texttt{++} further provides programmatically defined rewards. These tasks are a first step towards more open-ended human-agent interactions in which a human specifies a task using natural language and provides subsequent judgement about performance \citep{li2016dialogue, ammanabrolu2020motivate, team2021creating}.

We focus on training agents to solve these tasks using methods that can in principle apply to any task one hopes to perform on a digital device, and that have desirable data- and compute-scaling properties. We therefore turn to a straightforward combination of reinforcement learning (RL) \citep{sutton2018reinforcement} and behavioural cloning (BC) \citep{pomerleau1989alvinn}, the latter of which is aided by an alignment between human and agent action spaces (i.e. keyboard and mouse). This is a combination that was proposed at MiniWob's conception \citep{shi2017world}, but was not found at the time to produce high-scoring agents. Subsequent work consequently attempted to improve performance by giving agents access to DOM-specific actions \citep{liu2018reinforcement, gur2018learning, jia2019dom}, curricula~\cite{gur2018learning}, and through constrained exploration techniques that reduce the number of actions available at each step using curated guidance \citep{liu2018reinforcement}.
Revisiting the simple and scalable combination of imitation and reinforcement learning, we find that the primary missing factor in achieving high performance is simply the size of the human trajectory dataset used for behavioural cloning. Performance increases reliably with increasing amounts of human data, with continuing improvements observed for datasets of up to $400\times$ the size of those used in previous studies. Our results using this data surpass previous state-of-the-art performance by a wide margin, and are even able to achieve human-level mean performance across the task suite.

\section{Methods}

\subsection{MiniWob\texttt{++}}
MiniWob\texttt{++} is a suite of web-browser based tasks  introduced in \citet{liu2018reinforcement} (an extension of the earlier MiniWob task suite~\cite{shi2017world}). Tasks range from simple button clicking to complex form-filling, for example, to book a flight when given particular instructions (Fig.~\ref{fig:concept}a). Programmatic rewards are available for each task, permitting standard reinforcement learning techniques.

Previous work on MiniWob\texttt{++} has considered architectures that have access to DOM-specific actions \citep{liu2018reinforcement, gur2018learning, jia2019dom}, allowing agents to directly interact with a DOM element (even if it is not actually visible) as opposed to needing to use a mouse or keyboard to navigate to it. We choose instead to use only mouse and keyboard based actions. This simplifies the use of human behavioural data, and we further hypothesise that this interface will better transfer to computer control tasks without a compact DOM to interact with (and non web-browser based tasks). Finally, many MiniWob\texttt{++} tasks require clicking or dragging actions that cannot be achieved with DOM-element based actions (see examples in Fig.~\ref{fig:concept}b).

In common with previous MiniWob\texttt{++} studies \citep{liu2018reinforcement, gur2018learning, jia2019dom}, we give our agent access to an environment-provided dictionary of text strings that must be inputted to a given task's input fields (see Appendix Fig.~\ref{fig:observations} for an example). This helps to avoid the exploration problem of learning a generative text model from only sparse RL inputs. Interestingly, as we discuss in Sec.~\ref{sec:ablations}, the agent is able to achieve state-of-the-art performance even without this input. 

The MiniWob environment is realtime, which introduces technical and algorithmic complications. For example, there are no guarantees about how actions lead to observations temporally -- observations may be influenced by lag due to competition over a computer's resources. For the human participants providing demonstration data, we ensure sufficient resources on the server running the environment to minimize temporal jitter on the local machine, on which we sync observations and actions at 30Hz. For agents, any temporal jitter is sure to be different than that experienced by humans. Nevertheless, most MiniWob\texttt{++} tasks are relatively time insensitive, and our results demonstrate that the mismatch is not problematic in practice. Moreover, we found that actively manipulating the demonstration data to remove steps without actions (an operation that will only serve to exaggerate the differences in timings) gave the best results. As in previous studies, our human and agent scores are not discounted by time to completion. 

\subsection{Environment Interface}
Agents need a suitable interface for transmitting and receiving observations and actions if they are to use computers as humans do. The original MiniWob\texttt{++} task suite is supplied with a Selenium-based interface \citep{liu2018reinforcement}. We decided to implement an alternative environment stack designed to flexibly support any task that can be achieved in a web browser. This interface is optimised from the ground up for security, features, and performance~(Fig.~\ref{fig:concept}a).

\paragraph{Security.} We run the web browser (Google Chrome in our case) in a Sandbox2 \citepalias{sandbox2} container, which provides a chroot jail and restricts access to any system calls that could be used to compromise the host system. We further reroute all network traffic to a local host-side socket using a TCP proxy server. This gives us precise control over the content that can be accessed from within the browser. Not only are these security features important for safe agent-environment interactions, they also simplify recording demonstration data on publicly available systems such as Amazon Mechanical Turk, because neither humans nor agents should be able to use the browser itself to attack the host system or any network resources.

The original MiniWob\texttt{++} environment implementation accesses the internal browser state and issues control commands via Selenium. We instead interface directly with the Chrome DevTools Protocol (CDP) to retrieve browser-internal information such as the DOM tree and execute developer-provided JavaScript code on a page. We access the CDP via file descriptors that are passed to the browser at start-up time. We do this to minimize the attack surface and follow the security recommendations provided by the web driver authors \citepalias{webdriver}.

\paragraph{Features.} We want our agents to be able to interact with a standard web browser via the same controls used by human desktop users. To achieve this, our environment connects directly to an X11 server to input mouse and keyboard controls as well as retrieve the current frame buffer. This minimises domain shift between human and agent environment interactions, which is a challenge for the original Selenium interface. For example, mouse dragging actions are difficult to implement in Selenium. 

Working with X11 has other key advantages: (1) agents can potentially interact with the full browser (including tabs and the address bar), (2) recording first-person demonstration data at scale is easy due to close human and agent environment equivalence, and (3) we can render the context-sensitive system mouse cursor. 

We implemented our entire environment stack in C\texttt{++} to enable low latency interactions. This is particularly important when recording demonstrations, as accurate mouse cursor movements are difficult if the total input-latency is too high.

\begin{figure*}[t!]
    \centering
    \includegraphics[width=1.\textwidth]{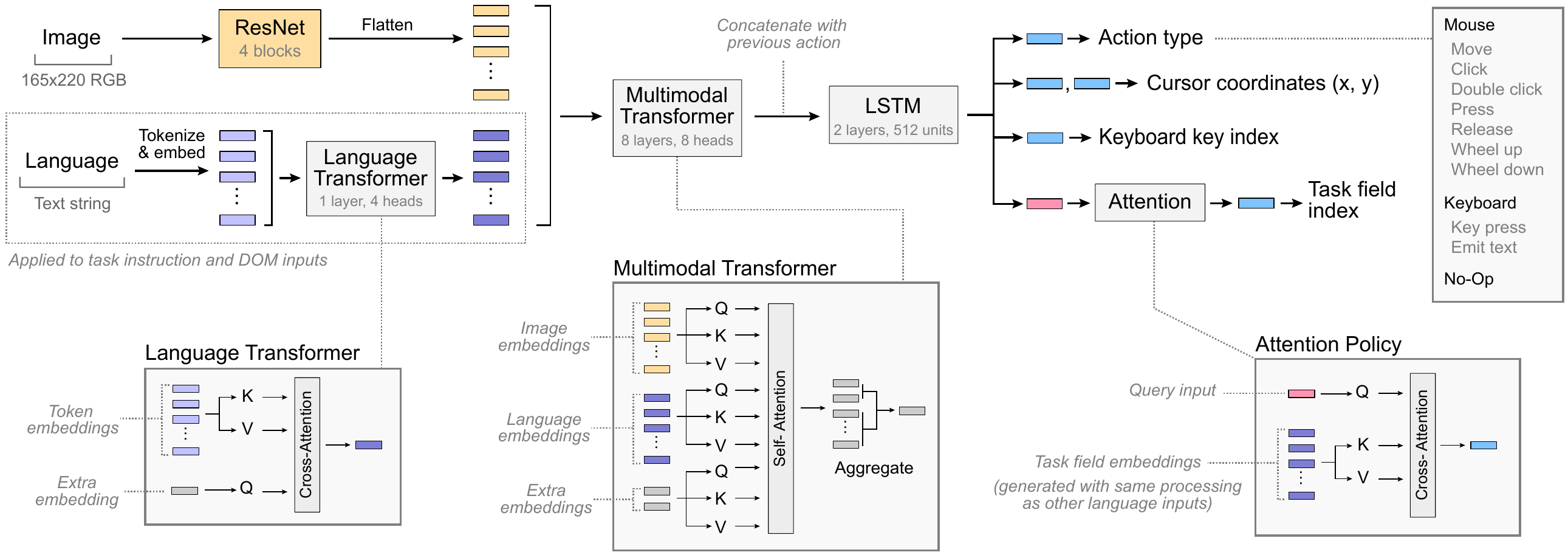}
    \vspace{-0.5cm}
    \caption{\textbf{The Computer Control agent architecture (CC-Net).} Pixel inputs are encoded by 4 ResNet blocks; language inputs (task instruction and DOM) are encoded via cross-attention. An 8-layer multimodal transformer combines pixel and language embeddings, whose output is concatenated with an embedding of the previous action and passed through 2 LSTMs. The agent produces 4 outputs: action type, cursor coordinates, keyboard-key index, and task-field index (which allows the agent to choose one of the task field strings to emit). All outputs are produced via linear transformations, with the exception of the task-field index, whose logits are produced via an attention-based policy.
}
    \label{fig:architecture}
\end{figure*}

\subsection{Agent Architecture}
\label{sec:agent_architecture}

Ultimately we found it to be unnecessary to implement specialised DOM-processing architectures based on, for example, graph nets~\cite{jia2019dom}. Instead, motivated by recent work on multimodal architectures, we applied minimal modality-specific processing, primarily relying on a multimodal transformer to flexibly attend to relevant information, as described below and in Fig.~\ref{fig:architecture}.

\paragraph{Perception.} The agent receives visual inputs (165x220 RGB pixels) and language inputs (an example input is shown in Appendix Fig.~\ref{fig:observations}). The pixel inputs pass through a series of four ResNet blocks, with $3\!\times\!3$ kernels, strides of $2$, $2$, $2$, $2$, and an increasing number of output channels ($32$, $128$, $256$, $512$). This results in $14\!\times\!11$ feature vectors, which we flatten into a list of $154$ tokens.

The three types of language inputs -- task instruction, DOM, and task fields (the last only used for the policy) -- are processed using the same module: each text string is split into tokens, with each token mapped to a size 64 embedding by indexing a learnable embedding table using a vocabulary of $370$ words. We reserve an additional $1000$ indexable embeddings for input words that fall outside the vocabulary. For these words, an index is calculated using a $64$-bit hash function with the output reduced to an integer between $370$ and $1369$. Using a 1-layer transformer with 4 heads (``language transformer'') we attend over the token embeddings of each individual string to produce a single 512-dimensional embedding using cross attention -- the token embeddings are used to generate keys and values, with an extra learnable embedding used as a query. The extra learnable embeddings provide input-independent components to the attention. This
is analogous to the special ``CLS'' tokens used in BERT
(Devlin et al, 2018) whose outputs can be directly used as
an aggregate output of the transformer.

\paragraph{Multimodal integration and memory.} The visual input embeddings, the language embeddings produced from DOM and task instructions, and two extra learned embeddings are fed into a multimodal transformer with 8 layers, 8 heads, and a 512 dimensional embedding.

The processed outputs corresponding to the extra embeddings are concatenated with the result of a feature-wise mean pooling operation across the remaining outputs and an embedding of the previous action. The resulting 1536-dimensional vector is input to a dual-layer LSTM with 512 hidden units per layer. Residual connections bypass each LSTM layer.

We have not run LSTM ablations in the final configuration, but earlier tuning showed that a dual-layer LSTM improved performance relative to a single-layer LSTM or to a feedforward architecture. It may be that this is due to the increased parameter count associated with the dual layer LSTM, as we did not control for this in our tuning. We note that there are some tasks for which memory is especially useful, such as those with moving elements.

\paragraph{Policy.} The agent's policy consists of 4 outputs: the action type, cursor coordinates, keyboard-key index and task-field index. Each output is modeled by a single discrete probability distribution, except for cursor coordinates which are modeled by two discrete distributions (for height and width coordinates), dividing each dimension into 51 bins. This is the only policy component that is auto-regressive, with the height coordinate conditioned on the width coordinate.

The action type is chosen from a set of 10 possible actions, which includes a \textit{no-op} (indicating no action), 7 mouse actions (move, click, double click, press, release, wheel up, wheel down), and two keyboard actions (key press, emit text). The \textit{key press} action is used to emit a keyboard key or one of a set of small macros (such as \texttt{CTRL+C}, full list in Appendix Table~\ref{tab:keys}). The \textit{emit text} action can be used to emit a string given by one of the MiniWob ``task-field'' observations.

Sampling of the remaining policy outputs is dependent on which action type was chosen -- i.e. cursor coordinates are sampled if \textit{move mouse} was chosen; a task-field index is sampled if \textit{emit text} was chosen (to determine which task-field string to emit); a keyboard-key index is sampled if \textit{key press} was chosen (to determine which key to emit).

The logits for the action-type, cursor coordinates and keyboard-key index distributions are produced via linear transformations, whereas the logits for the task-field index are produced via dot-product attention. This attention policy works as follows: we first use a linear output to produce a query. A separate linear transformation is used to produce keys from the embeddings of the available task fields. The logits are the attention weights that result from the dot-product between query and keys, followed by a softmax.

In the ablation experiments in Section~\ref{sec:ablations}, we use two alternative action types which allow the agent to act directly on the DOM -- \textit{DOM click}, for triggering a click on a specific DOM element, and \textit{DOM emit text}, for directly writing a task-field string into a specified DOM element (i.e. in this macro-action a DOM element is focused on and then text is emitted into it). These actions require an additional policy output (DOM element index) to determine which DOM element to click. This was produced using an attention policy as described above, but with attention over DOM element embeddings. For the  \textit{DOM emit text} action, we reused the task-field index output to determine which text string to emit.

\subsection{Human Data Collection}
\label{sec:human_data}

Given that agents and humans use the same interface, the use of human demonstrations was considered in the original MiniWob publication \citep{shi2017world}. However, only $17$ hours of human data were collected for that study (with pure behavioural cloning policies only solving $5\%$ more tasks compared to a random policy), with a further $1000$ demonstrations per task for a handful of tasks utilised in the later MiniWob\texttt{++} paper \citep{liu2018reinforcement}. Given recent trends that indicate the effect of data scaling on performance \citep[e.g.]{team2021creating,kaplan2020scaling}, it is natural to revisit the use of human demonstrations in larger quantities. 

We collected over 2.4 million demonstrations of the $104$ MiniWob\texttt{++} tasks from a total of $77$ human participants, which amounts to approximately $6300$ hours. Participants were recruited via a crowdsourcing data collection platform and paid a fixed hourly rate. Human data-collection was subject to an ethical review process.

Humans can perform many of the MiniWob\texttt{++} tasks well without practice. However a subset require practice, learning of specialized knowledge (e.g. POSIX terminal commands), understanding small amounts of ambiguity in the intent of the templated goal, or adjusting to timing or lag. Accordingly, we observed human performance increased as they practiced (data not shown), and as described below, unsuccessful trajectories were filtered out of the dataset.

\subsection{Training}

We trained agents using a straightforward mixture of imitation learning (behavioural cloning (BC) \citep{pomerleau1989alvinn}) and reinforcement learning (RL) (using the VMPO algorithm \citep{song2019v}), which is a combination that has been largely successful in many challenging domains \citep{silver2016mastering,vinyals2019grandmaster,stiennon2020learning,liu2018reinforcement,shi2017world}. Training hyper-parameters are provided in Appendix Table~\ref{tab:hparams}.

To combine imitation and reinforcement learning one can: (1) Co-train from scratch with a weighted mixture of the two losses; (2) Pre-train using BC and subsequently tune with RL, with or without divergence penalties back to the BC policy \citep[e.g.]{vinyals2019grandmaster}. In our work we consider the former with equal weighting to the BC and RL losses, though we have early evidence that pre-training with BC may improve the efficiency of learning (data not shown).

Before being used for BC, our human demonstration data was split into train and test sets (2.2 million \& 310 thousand episodes respectively). These episodes were further filtered by success; that is, tasks where the final reward was less than $0.5$ were removed ($\sim\!5\%$). Additionally, we cleave ``no-op'' steps from the demonstrations, which are steps where humans did not act, up to a maximum of $10$ consecutive steps. Agents were trained indiscriminately on this data: all tasks were co-trained, using uniform sampling across demonstrations, which might introduce slight asymmetries across task representations depending on the data collected, the filtering procedure, and the length of the demonstrations.

We choose to co-train on all tasks for two reasons. First, we find that there is a significant transfer effect, with training being more efficient per frame of each task seen for co-training (see Sec.~\ref{sec:task_transfer}). Second, our ultimate objective is a generally useful agent, and we therefore require one agent with as many capabilities as possible.  

Finally, due to the computational requirements of our agents, tuning was done by hand based on a limited number of experiments. We were furthermore only able to run one seed for each of the conditions that we report in this paper.

\begin{figure}[t!]
    \centering 
    \includegraphics[width=.48\textwidth]{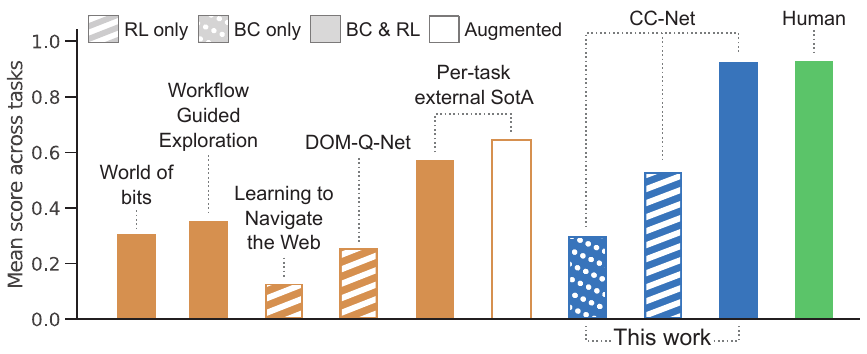}
    \vspace{-0.5cm}
    \caption{\textbf{Mean performance across MiniWob\texttt{++} tasks.} Shown are previously published MiniWob\texttt{++} results \citep{shi2017world, liu2018reinforcement, gur2018learning, jia2019dom}, which each only report performance for a subset of tasks (for these results the performance on non-reported tasks is set to zero, all scores used are given in Appendix Table~\ref{tab:all_scores}). In order to make a fair comparison, we combine the best external results for each task into aggregated external SotA scores, reporting those obtained both with and without additional augmentations (we filter the 16 out of 104 tasks without external scores from our corresponding results). Our agent, CC-Net, outperforms these results by a large margin, and matches mean human performance. We achieve this without using environment augmentations. Instead, we find that the combination of BC and RL is crucial to CC-Net performance; BC- and RL-only training are less effective.}
    \label{fig:core_results}
\end{figure}

\begin{figure*}[t!]
    \centering 
    \includegraphics[width=1.\textwidth]{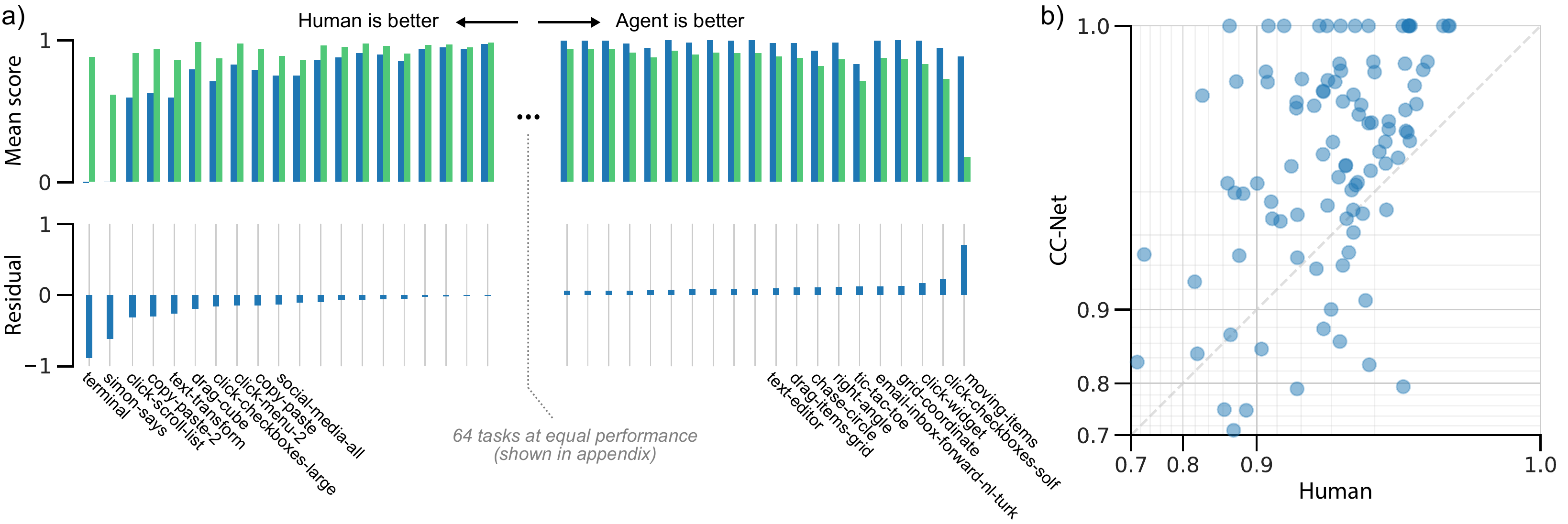}
    \vspace{-0.7cm}
    \caption{\textbf{Per-task comparison of CC-Net and human performance.} a) On a small number of tasks humans (green) perform better than our agent (blue), and vice versa. For the large majority, performance differences are minimal, as both humans and agents solve the tasks nearly perfectly (see Appendix Fig.~\ref{fig:per_level_vs_human_all} for all tasks). b) Depicted in the scatter-plot on the right are performance scores for all 104 tasks save for three outliers (the polar extremes of the bar plot where one of human or agent scores are $<0.7$).}
    \label{fig:per_level_vs_human}
\end{figure*}

\begin{figure}[t!]
    \centering 
    \includegraphics[width=.42\textwidth]{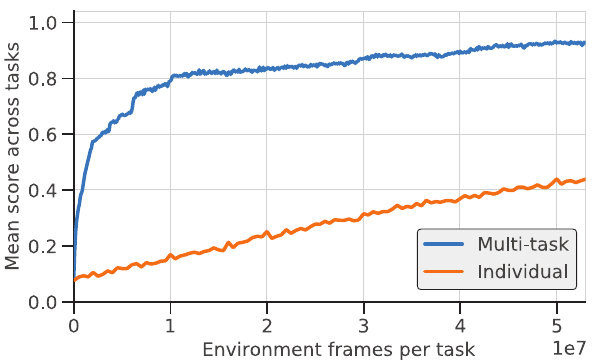}
    \vspace{-0.2cm}
    \caption{\textbf{Comparison between single and multi-task training.} Co-training on all 104 MiniWob tasks is significantly more data efficient than training on each task individually, providing evidence of transfer between tasks. }
    \label{fig:multi_task}
\end{figure}

\begin{figure*}[h!]
    \centering
    \includegraphics[width=1.0\textwidth]{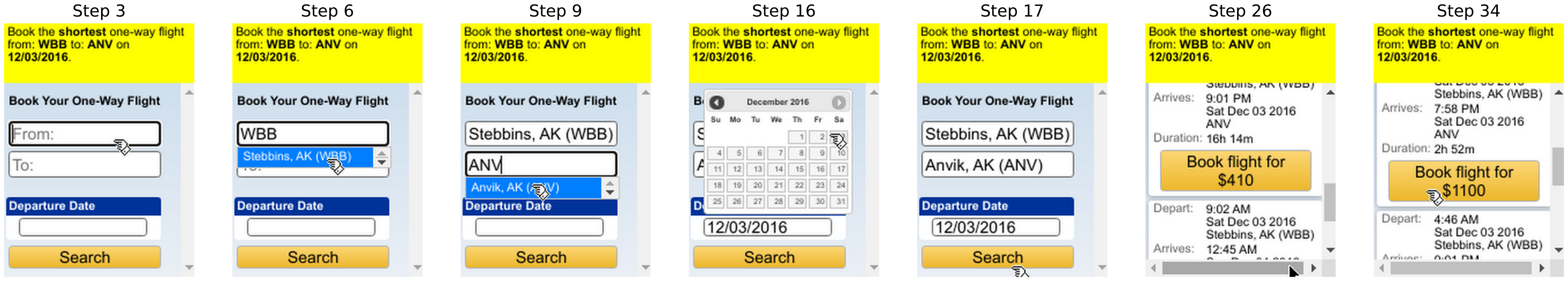}
    \vspace{-0.5cm}
    \caption{\textbf{Selected steps from a successful agent trajectory on \texttt{book-flight}.} This task has over $1.9$ billion possible permutations, representing a significant exploration challenge for pure reinforcement learning. We find that the combination of BC \& RL is sufficient to solve it, demonstrating the utility of using human data to shape behavioural priors.}
    \label{fig:book_flight_episode}
\end{figure*}

\section{Results}

\subsection{Human-Level Performance on MiniWob\texttt{++}}  

It is challenging to directly compare our performance to the previous literature, as papers have typically only tackled a subset of the MiniWob\texttt{++} tasks. We therefore take the best published performance on each individual task, and use this aggregated performance as a comparison for our agent (we perform separate aggregations for both with and without curricula or other augmentations). Our agent exceeds this SotA benchmark performance by a wide margin (Fig.~\ref{fig:core_results}, a per-task comparison to external SotA is reported in Appendix Fig.~\ref{fig:per_level_vs_sota}). In addition, we find that our agent achieves human-level mean performance across the suite of MiniWob\texttt{++} tasks. This performance is enabled by our combination of BC and RL co-training -- as shown in Fig.~\ref{fig:core_results}, ablations without both training signals perform much worse. Fig.~\ref{fig:book_flight_episode} shows frames of a successful agent trajectory on the challenging task \texttt{book-flight}.

Examining our performance across the task suite in detail, we find that, while our mean score matches mean human performance, there are tasks for which humans significantly out-perform our agent (Fig.~\ref{fig:per_level_vs_human}). In particular, our agent fails to get any score for two tasks: \texttt{simon-says}, and \texttt{terminal}. \texttt{simon-says} is challenging even for humans. It requires remembering a random sequence of button pushes, and then repeating this sequence back. Our agent observes the environment at about 2Hz, meaning that it frequently misses the presentation of at least one button in the sequence. \texttt{terminal} involves using a Unix terminal to search for and delete a file. Human participants required training to perform this task before they achieved high performance, as most had no experience with a Unix terminal. Further investigation is needed to understand why our agent fails to get any score on this task, especially given that it is able to solve related key-pressing tasks such as \texttt{text-transform}, \texttt{simple-arithmetic} and \texttt{enter-text-2}
(see Appendix Fig.~\ref{fig:text_transform_histogram} for the histogram of key presses for 100 episodes of \texttt{text-transform}). There are a few tasks for which our agent performs significantly above mean human performance. The biggest performance gap is for the \texttt{moving-items} task where human participants rarely succeeded at all, most likely due to network-related control latency.

\subsection{Task Transfer}
\label{sec:task_transfer}

We find that, with a fixed budget of training steps per task, training one agent on all 104 MiniWob\texttt{++} tasks leads to a strong boost in performance compared to training separate agents on each task (Fig.~\ref{fig:multi_task}).
In order to compare these training regimes, we measured performance for each task expressed as a function of frames consumed on that individual task. The plot shows the average across all tasks. These results provide a promising hint of the potential of such a unified control interface to support generalisation across a diverse range of human tasks. Similar results were found on a smaller scale (training on 9 tasks simultaneously) in \citet{jia2019dom}.

\subsection{Scaling}

\begin{figure}[t!]
    \centering
    \includegraphics[width=0.4\textwidth]{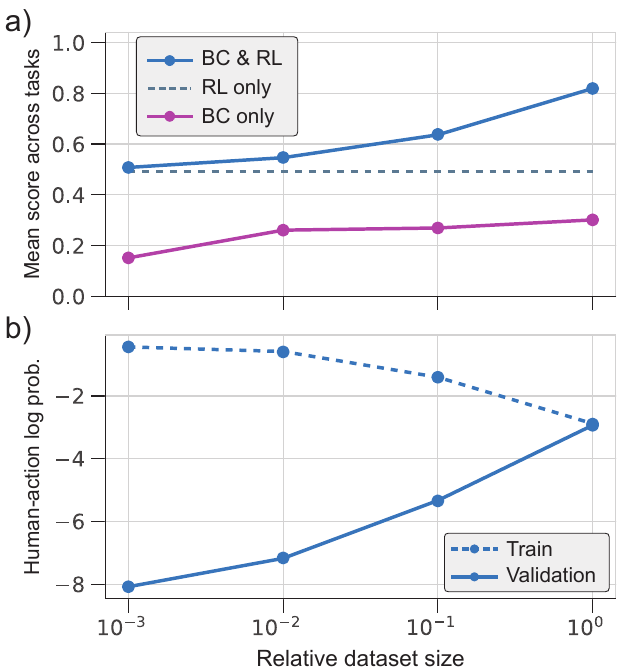}
    \vspace{-0.1cm}
    \caption{\textbf{Agent performance for varying dataset sizes}: a) Performance at 2 billion frames for BC \& RL co-training using different dataset sizes. Also shown are maximum BC-only training performances. b) Corresponding log probabilities for agent predictions of human actions on train (solid) and validation (dashed) trajectories under BC \& RL co-training. As can be seen, significant overfitting occurs for smaller dataset sizes.}
    \label{fig:dataset_scaling}
\end{figure}

The size of the human trajectory dataset is a crucial factor in agent performance (Fig.~\ref{fig:dataset_scaling}). Using 1/1000 of the dataset, which roughly corresponds to 6 hours of data, leads to rapid overfitting and no significant gain over RL only performance. We see continuing performance gains as we increase the amount of data from this baseline over three orders of magnitude up to the full dataset size. We note that higher performance at these dataset sizes could be possible with changes in algorithm or architecture. For example, as we reduced the dataset size, we simply reduced the relative weight of the BC loss (scaled by the square root of the relative size) to slow overfitting, when it may in fact be more effective to instead use a KL divergence penalty to a pre-trained BC-only policy that is trained to the point of minimum validation loss. Nevertheless, the research and experimentation costs required to explore avenues that afford efficiency gains in low data-regimes should be balanced with the simplicity of collecting more data.

\subsection{Input and Output Ablations}
\label{sec:ablations}

Our agent consumes both pixel and DOM information, and can be configured to support a range of different possible actions. We perform ablations to understand the importance of these different architectural choices.

\begin{figure}[t!]
    \centering
    \includegraphics[width=0.44\textwidth]{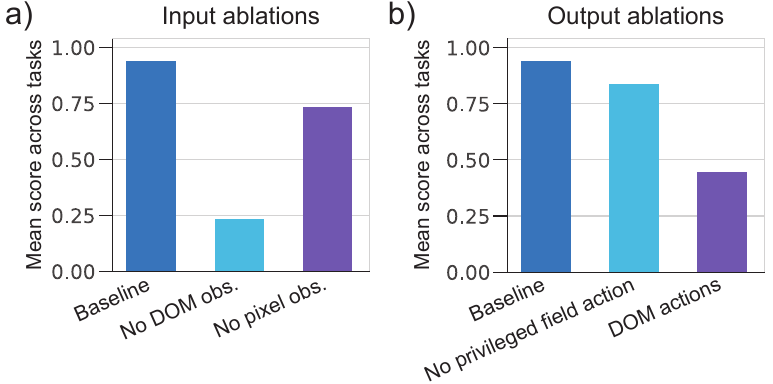}
    \vspace{-0.2cm}
    \caption{\textbf{Ablations.} a) Agent input ablations: Removing the DOM input observation leads to a large drop in performance, while removing the pixel observation is less deleterious. b) Output ablations: The agent is surprisingly effective without text input choices (task fields) provided by the environment. This is because it is able to learn to use the mouse to copy the text from the prompt into the required input field. Using an action space in which the agent interacts with specified DOM elements directly is less effective than our baseline mouse and keyboard actions.}
    \label{fig:ablations}
\end{figure}

We first ablate the different agent inputs (Fig.~\ref{fig:ablations}a). Our current agent configuration is strongly reliant on DOM information, with a 75\% drop in performance if this input is removed. In contrast, the agent is less sensitive to operating without visual information. This is even the case for tasks that require processing of a ``canvas" with shapes and lines. An inspection of the DOM shows that, for MiniWob\texttt{++}, this canvas is cleanly represented within the DOM, making such information available to the agent even without pixel information. This would not be the case for more general computer control tasks, pointing to the importance of improving our agent's pixel-only performance. 

In Fig.~\ref{fig:ablations}b we show an ablation in which we remove the agent's ability to use the text input choices (task fields) given by the environment. Interestingly, such an agent is still able to solve tasks involving form filling, however, it learns from human trajectories to do this by highlighting text, and then dragging this to the relevant text box. Notably, this dragging action would not be straightforward for agents to achieve in the original Selenium version of the environment.

We further show an ablation in Fig.~\ref{fig:ablations}b in which the agent uses alternative actions that interact with a specified DOM element (described in Section~\ref{sec:agent_architecture}). This means that the agent is unable to solve tasks that involve clicking at a specific location within a canvas, dragging, or highlighting. As we did not optimise for this agent architecture, there is a significant mismatch between our recorded human trajectories and the actions available to the agent. During imitation learning, we simply ignore these actions, but this nonetheless leads to the agents being nudged towards trajectories that are not necessarily optimal for DOM-based actions. In an additional experiment, we found that augmenting our original action set with DOM-based actions did not lead to a boost in performance over the baseline (data not shown).

\section{Discussion, Limitations, \& Related Work}

\paragraph{Device assistance.} Creating agents that effectively assist us on our digital devices requires: (1) understanding human intent, language, and judgement, and (2) the ability to turn language-specified goals into actions on a device. The MiniWob\texttt{++} benchmark \citep{liu2018reinforcement} serves as a jumping off point for studying the second requirement: agents must turn language goals into actions for a modern web browser. While we show that deep reinforcement learning, augmented with human data, is well positioned for solving this issue, the MiniWob\texttt{++} benchmark hides the complexity of the first requirement; work remains to tackle (1) and (2) in concert to produce useful agents on our devices.

MiniWob\texttt{++} prescribes goals via natural (but scripted) language that is precisely mapped to reward outcomes by code, putting aside the difficulties of dealing with human intent, language, and preferences. Reward information is thus low-noise, undelayed, and available in large quantities. Recent work has studied how to align RL agents with the complexity of human preferences and human language. For example, \citet{christiano2017deep} trained agents via reinforcement learning and using reward grounded in human feedback to play Atari and perform motor control tasks. Interest in human alignment has accelerated \cite{wirth2017survey,ibarz2018reward}, and spread to a variety of domains, including dialogue agents \cite{jaques2019way}. In parallel, large-scale language models built on the Transformer architecture \cite{vaswani2017attention} have revolutionized our capacity to build agents that consume and generate natural language. GPT-3 \cite{brown2020language} and related work \cite{shoeybi2019megatron, rae2021scaling} has shown that pre-training on vast quantities of human generated text creates models that are fast and flexible learners for a dizzying array of language understanding tasks. A recent language model called Codex \cite{chen2021evaluating} has shown the potential of these ``foundation models'' \cite{bommasani2021opportunities} for connecting natural language intent with our computers by fine-tuning models on publicly available code. 

A growing body of work has begun to connect these pieces: \citet{ammanabrolu2020motivate} build agents that both act and communicate in pursuit of goals in a crowd-sourced fantasy text-game. They incorporate large language models and human demonstrations to derive agents that can act and speak to effectively pursue goals. \citet{team2021creating} use human demonstrations and judgement to construct agents that are responsive to human language and intent in a 3D game environment. Most relevant here, WebGPT \cite{nakano2021webgpt} builds on large-scale language modelling results to create a browser-assisted question-answering agent that is enhanced by human feedback. This agent acts on the internet using a bespoke text interface which allows it to search, follow links, and quote sources. By using the internet, the agent improves its question-answering capacities, as judged by human participants.

\paragraph{Data.} Approximately $400\times$ more data than was initially collected \cite{shi2017world} is required for achieving human-level performance. While this is a lot in comparison to the initial collection, it is also a \textit{one-time} cost, and is orders of magnitude smaller than recent datasets used for language modeling, for example. Moreover, the costs associated with the experimentation of new algorithms and architectures that work well on lower amounts of data can very quickly swamp those required for simply collecting more data.

One might worry that our dataset exhausts every possible permutation of MiniWob\texttt{++} task presentation, rendering the task of ``solving'' tasks to simple memorization. However, this is far from true since the possible permutations of task-presentation vastly exceeds the amount of data we collected for it. Moreover, there is natural variance embedded in the environment, including, for example, initial mouse cursor location. Finally, the significant difference between individual versus multi-task training suggests that there are many tasks for which transferable skills can be gained from other tasks, which would likely not be the case if each task was solved by simply memorising its required outputs.

\paragraph{Performance.} There are a number of unexplored approaches that might improve performance. For example, we could skew the presentation of behavioural cloning data so that harder tasks are over-represented. Or, we could tune the RL and BC objectives, perhaps using BC pre-trainined and a decaying KL-penalty back towards the BC policy during RL \citep{vinyals2019grandmaster}. Nevertheless, part of the goal of this work was to show the effectiveness of a straightforward technique, executed with minimal bells-and-whistles. 

While our approach does not lie on the cutting edge of algorithmic novelty, it provides a resolution to a concrete hypothesis introduced in the original MiniWob work; namely, the effectiveness of deep RL methods for solving tasks on a computer involving keyboards and mice. Indeed, examining the literature in this space would suggest that conventional approaches are inadequate, and more specialized techniques are needed. We have shown this to not be the case, so long as one appreciates the role of demonstration data in augmenting standard deep RL techniques.

\paragraph{Conclusion.} Humans use digital devices for billions of hours every day. If we can develop agents that can assist with even a tiny fraction of these tasks, we can hope to enter a virtuous cycle of agent assistance, followed by human feedback on failures, and hence to agent improvement and new capabilities. 

\section{Contributions}
AS, TP, TL and PCH conceived the project. AG managed the project. GT, AS, TP, AM, PCH, PG, RC and JA developed the environment, data collection, experiment and agent infrastructure. AS, DR and PCH developed the agent architecture, ran experiments and analysed data. AS, DR, TL and PCH wrote the manuscript.

\section{Acknowledgements}

We thank the DeepMind Interactive Agents Team~\citep{team2021creating} for helping us to use their infrastructure as the base for our experiments. We also thank the DeepMind Platform Team and Crowd Compute for their work contributing to our human data collection pipeline and environment.

\bibliographystyle{abbrvnat}
\bibliography{main}

\newpage
\appendix
\onecolumn

\begin{figure*}
    \centering
    \includegraphics[width=1.0\textwidth]{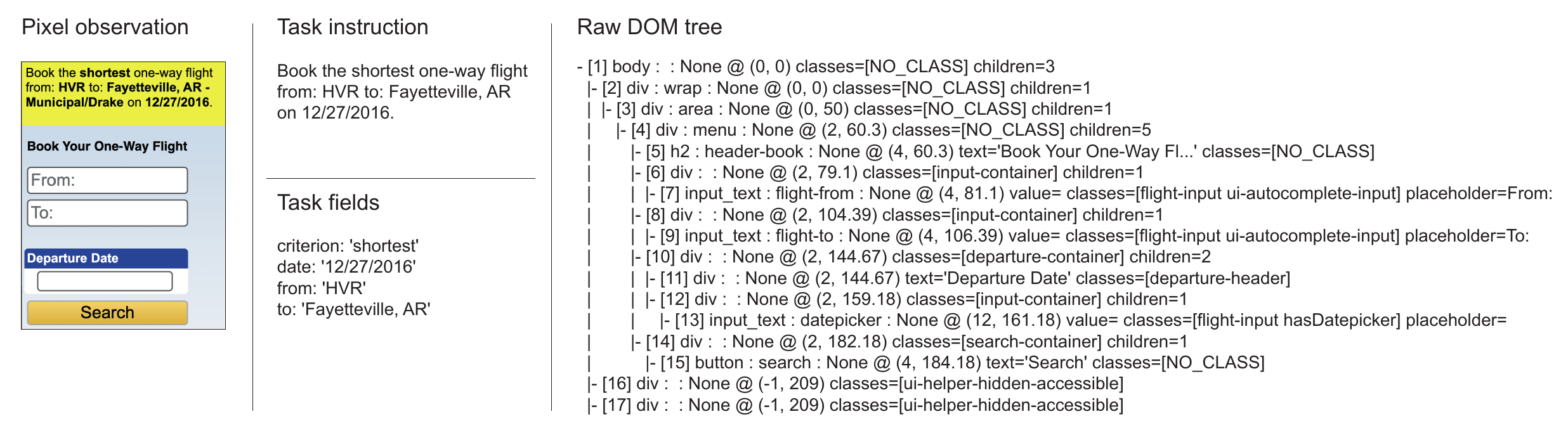}
    \caption{An example of the MiniWob observations for the \texttt{book-flight} task. The raw DOM tree is processed into a list of DOM elements. The information included in this list for each DOM element includes an integer reference, its text, any input text, the class, the state of e.g. radio buttons, its position, whether it is focused, and if it has been interacted with in the episode so far.}
    \label{fig:observations}
\end{figure*}

\begin{table}
\centering
\caption{Hyper-parameters used in training. See \citet{song2019v} for descriptions of the VMPO hyper-parameters.}
\vspace{0.2cm}
\small{
\begin{tabular}{|l|l|}
\hline
\sc{Parameter} & \sc{Value} \\
\hline
Optimizer & Adam~\citep{kingma2014adam} \\
Learning rate & $1\mathrm{e}{-4}$ \\
Adam $b_1$ parameter & 0.9 \\
Adam $b_2$ parameter & 0.999 \\
Weight decay (biases excluded) & $1\mathrm{e}{-1}$ \\
VMPO loss weight & 1.0 \\
BC loss weight (baseline) & 1.0 \\
VMPO $\epsilon_\alpha$ & 0.1 \\
VMPO $\epsilon_\eta$ & 0.2 \\
Agent discount $\gamma$ & 0.9 \\
Batch size & 256 \\
Trajectory unroll length & 64 \\
Target-network update period $T$ & 50 \\
Maximum number of steps per episode & 300 \\
\hline
\end{tabular}
}
\label{tab:hparams}
\end{table}

\begin{table}
\centering
\caption{List of keyboard keys and macros that the agent can emit in a \textit{key press} action.}
\vspace{0.2cm}
\small{
\begin{tabular}{|l|l|l|l|}
\hline
Enter & Digit6 & KeyG & ControlRight+KeyV \\
PageUp & Digit7 & KeyH & ShiftLeft+KeyA \\
PageDown & Digit8 & KeyI & ShiftLeft+KeyB \\
Backspace & Digit9 & KeyJ & ShiftLeft+KeyC \\
Delete & Digit0 & KeyK & ShiftLeft+KeyD \\
Tab & Numpad0 & KeyL & ShiftLeft+KeyE \\
Space & Numpad1 & KeyM & ShiftLeft+KeyF \\
ArrowUp & Numpad2 & KeyN & ShiftLeft+KeyG \\
ArrowRight & Numpad3 & KeyO & ShiftLeft+KeyH \\
ArrowDown & Numpad4 & KeyP & ShiftLeft+KeyI \\
ArrowLeft & Numpad5 & KeyQ & ShiftLeft+KeyJ \\
BracketLeft & Numpad6 & KeyR & ShiftLeft+KeyK \\
BracketRight & Numpad7 & KeyS & ShiftLeft+KeyL \\
Minus & Numpad8 & KeyT & ShiftLeft+KeyM \\
Equal & Numpad9 & KeyU & ShiftLeft+KeyN \\
Semicolon & NumpadAdd & KeyV & ShiftLeft+KeyO \\
Quote & NumpadMultiply & KeyW & ShiftLeft+KeyP \\
Backslash & NumpadSubtract & KeyX & ShiftLeft+KeyQ \\
Comma & NumpadDivide & KeyY & ShiftLeft+KeyR \\
Period & NumpadDecimal & KeyZ & ShiftLeft+KeyS \\
Slash & NumpadEnter & ControlLeft+KeyA & ShiftLeft+KeyT \\
Backquote & KeyA & ControlRight+KeyA & ShiftLeft+KeyU \\
Digit1 & KeyB & ControlLeft+KeyC & ShiftLeft+KeyV \\
Digit2 & KeyC & ControlRight+KeyC & ShiftLeft+KeyW \\
Digit3 & KeyD & ControlLeft+KeyX & ShiftLeft+KeyX \\
Digit4 & KeyE & ControlRight+KeyX & ShiftLeft+KeyY \\
Digit5 & KeyF & ControlLeft+KeyV & ShiftLeft+KeyZ \\
\hline
\end{tabular}
}
\label{tab:keys}
\end{table}

\begin{figure*}
    \centering
    \includegraphics[width=1.0\textwidth]{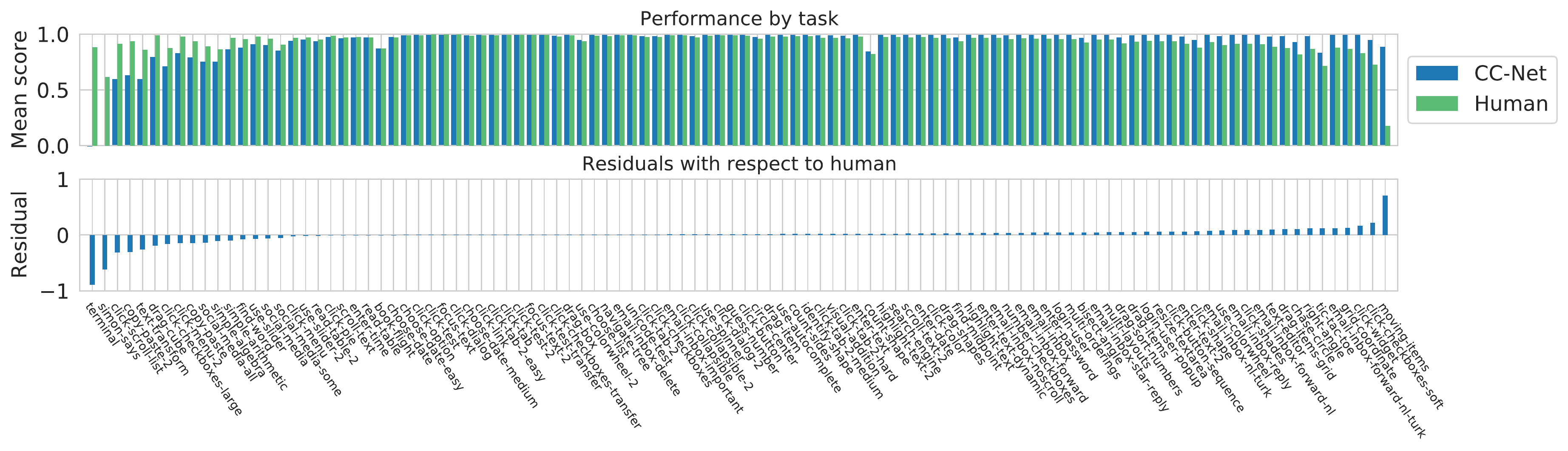}
    \caption{Per-task performance comparison to human participants. An expanded version of the plot in Fig.~\ref{fig:per_level_vs_human}a.}
    \label{fig:per_level_vs_human_all}
\end{figure*}

\begin{figure*}
    \centering
    \includegraphics[width=1.0\textwidth]{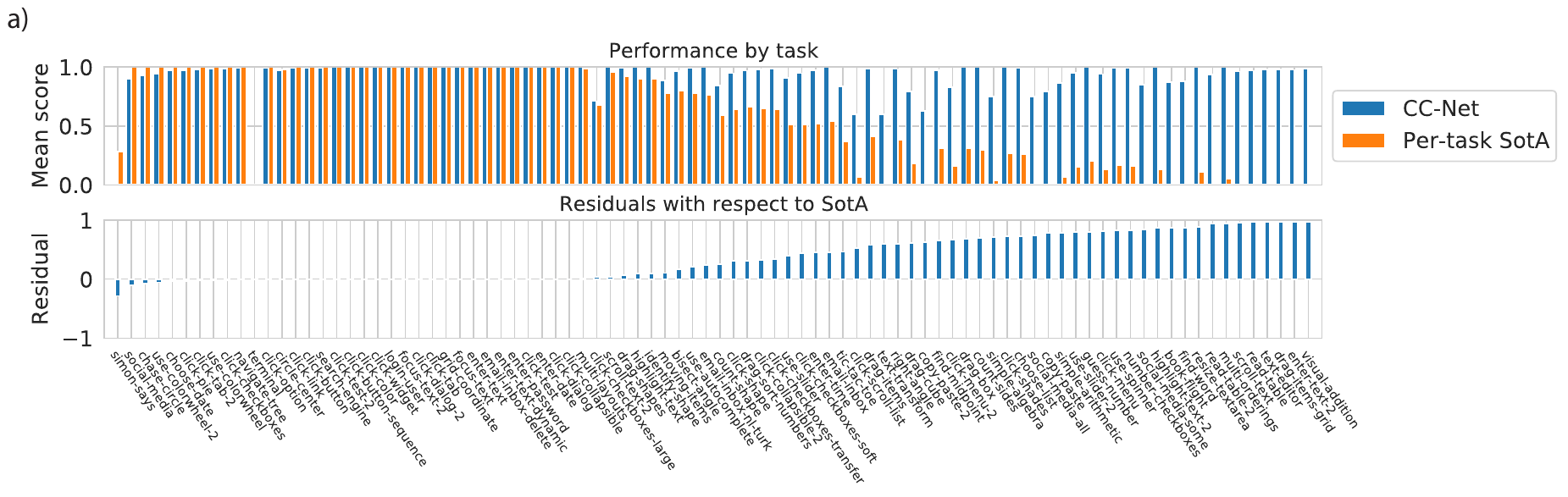}
    \caption{Per-task performance comparison to the best published external SotA results for each task. These SotA scores correspond to the aggregated scores shown in Table~\ref{tab:all_scores}.}
    \label{fig:per_level_vs_sota}
\end{figure*}

\begin{figure*}
    \centering
    \includegraphics[width=1.0\textwidth]{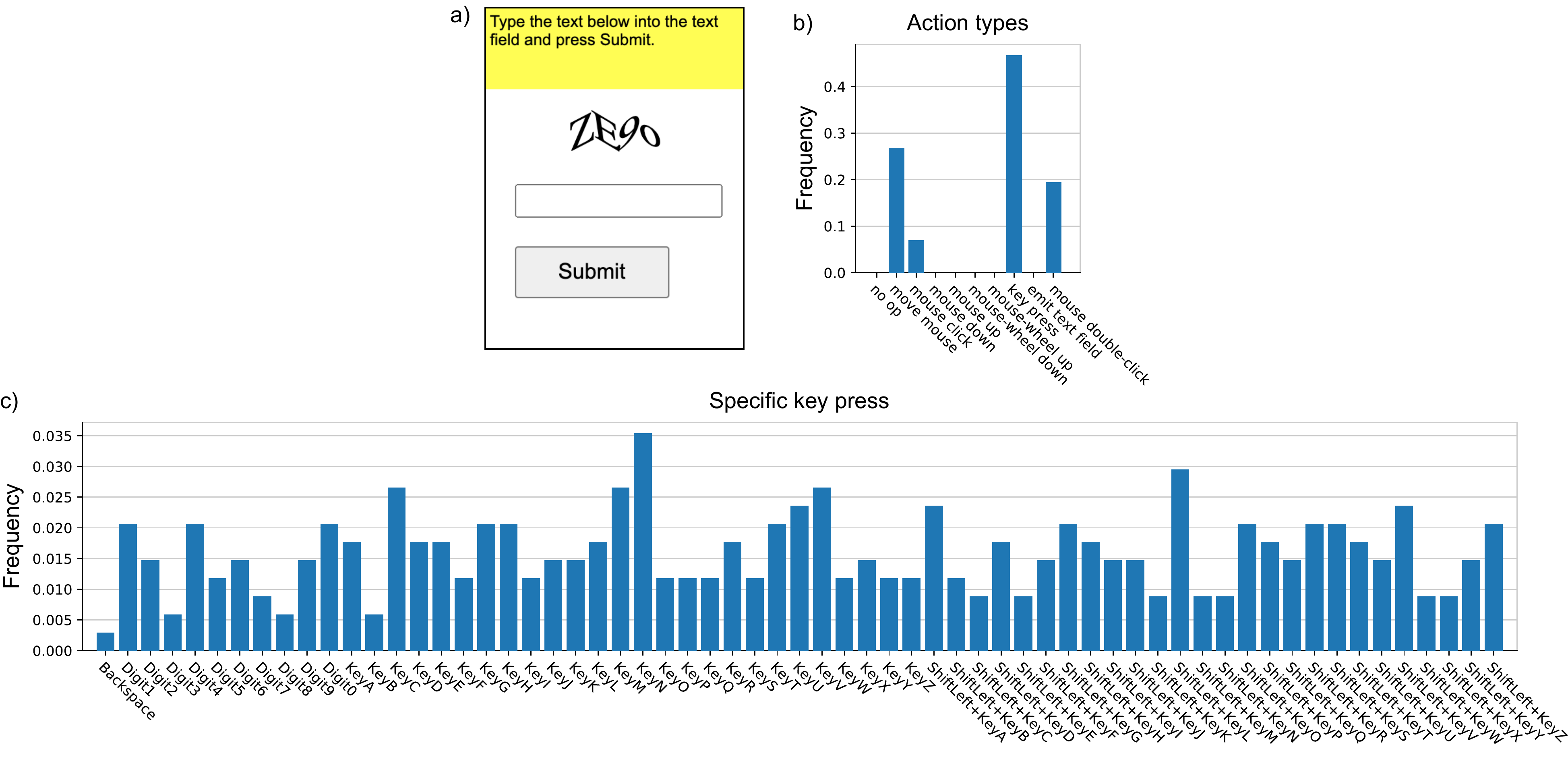}
    \caption{a) Initial frame of an example episode of \texttt{text-transform}. Note that the text (\textit{captcha}) to be typed is also present in the DOM, and thus passed to the agent as a language input. b) Distribution of action types selected by the agent for a random sample of $100$ episodes of \texttt{text-transform}. In this task the agent most frequently chooses to output a key press. c) Distribution of specific key-press choices for the episodes included in b). The agent learns to produce individual key presses for every letter (a-z, lower and upper case) and all digits (0-9), in order to complete the \textit{captcha}.}
    \label{fig:text_transform_histogram}
\end{figure*}

\clearpage
\begin{tiny}
\begin{longtable}{| p{2.4cm} *{11}{ | p{0.8cm}} |}
\caption{Per-task mean scores for human participants, our agent (CC-Net) and scores gathered from external studies~\citep{shi2017world, liu2018reinforcement, gur2018learning, jia2019dom}. Some scores from these studies were only available in figure format and were therefore estimated as accurately as possible. The final two columns give the best externally reported score for each task using either BC \& RL, or BC, RL and augmentations.}
\label{tab:all_scores}\\
\hline 
\sc{Task} \rule{0pt}{1.2\normalbaselineskip} &
Human & CC-Net & CC-Net BC only & World of bits (BC \& RL) & Workflow guided exploration (BC \& RL) & Learning to navigate the web (RL) & DOM-Q-Net (RL) & Workflow guided exploration (Augmented) & Learning to navigate the web (Augmented) & Aggregated SotA (BC \& RL) & Aggregated SotA (Augmented)\\
\hline
bisect-angle & 0.92 & 0.97 & 0.29 & 0.80 & n/a & n/a & n/a & n/a & n/a & 0.80 & 0.80 \\
book-flight & 0.87 & 0.87 & 0.00 & 0.00 & 0.00 & n/a & n/a & 0.00 & 1.00 & 0.00 & 1.00 \\
chase-circle & 0.82 & 0.93 & 0.80 & 1.00 & n/a & n/a & n/a & n/a & n/a & 1.00 & 1.00 \\
choose-date-easy & 0.99 & 0.99 & 0.42 & n/a & n/a & n/a & n/a & n/a & n/a & n/a & n/a \\
choose-date-medium & 0.98 & 0.99 & 0.26 & n/a & n/a & n/a & n/a & n/a & n/a & n/a & n/a \\
choose-date & 0.97 & 0.97 & 0.12 & 0.00 & 0.00 & n/a & 1.00 & 0.00 & n/a & 1.00 & 1.00 \\
choose-list & 0.98 & 0.99 & 0.19 & 0.25 & 0.16 & 0.26 & n/a & 0.16 & 0.26 & 0.26 & 0.26 \\
circle-center & 0.96 & 0.97 & 0.36 & 0.98 & n/a & n/a & n/a & n/a & n/a & 0.98 & 0.98 \\
click-button-sequence & 0.94 & 1.00 & 0.47 & 0.22 & 0.99 & n/a & 1.00 & 1.00 & n/a & 1.00 & 1.00 \\
click-button & 0.98 & 1.00 & 0.78 & 0.62 & 1.00 & 1.00 & 1.00 & 1.00 & 1.00 & 1.00 & 1.00 \\
click-checkboxes-large & 0.87 & 0.71 & 0.00 & n/a & 0.68 & n/a & n/a & 0.84 & n/a & 0.68 & 0.84 \\
click-checkboxes-soft & 0.73 & 0.95 & 0.04 & n/a & 0.51 & n/a & n/a & 0.94 & n/a & 0.51 & 0.94 \\
click-checkboxes-transfer & 0.98 & 0.99 & 0.36 & n/a & 0.64 & n/a & n/a & 0.64 & n/a & 0.64 & 0.64 \\
click-checkboxes & 0.97 & 0.98 & 0.32 & 0.48 & 0.98 & n/a & 1.00 & 1.00 & n/a & 1.00 & 1.00 \\
click-collapsible-2 & 0.97 & 0.98 & 0.17 & 0.11 & 0.65 & n/a & n/a & 0.99 & n/a & 0.65 & 0.99 \\
click-collapsible & 0.99 & 1.00 & 0.81 & 0.98 & 1.00 & 1.00 & n/a & 1.00 & 1.00 & 1.00 & 1.00 \\
click-color & 0.97 & 1.00 & 0.82 & 0.23 & 1.00 & n/a & n/a & 1.00 & n/a & 1.00 & 1.00 \\
click-dialog-2 & 0.99 & 1.00 & 0.88 & 0.53 & 1.00 & n/a & n/a & 1.00 & n/a & 1.00 & 1.00 \\
click-dialog & 1.00 & 1.00 & 0.95 & 1.00 & 1.00 & 1.00 & 1.00 & 1.00 & 1.00 & 1.00 & 1.00 \\
click-link & 0.99 & 0.99 & 0.59 & 0.31 & 1.00 & 1.00 & 1.00 & 1.00 & 1.00 & 1.00 & 1.00 \\
click-menu-2 & 0.98 & 0.83 & 0.52 & 0.16 & n/a & n/a & n/a & n/a & n/a & 0.16 & 0.16 \\
click-menu & 0.97 & 0.94 & 0.22 & 0.13 & n/a & n/a & n/a & n/a & n/a & 0.13 & 0.13 \\
click-option & 0.99 & 0.99 & 0.21 & 0.28 & 1.00 & n/a & 1.00 & 1.00 & n/a & 1.00 & 1.00 \\
click-pie & 0.98 & 0.97 & 0.15 & 0.15 & 0.32 & 1.00 & n/a & 0.32 & 1.00 & 1.00 & 1.00 \\
click-scroll-list & 0.91 & 0.60 & 0.01 & 0.07 & n/a & n/a & n/a & n/a & n/a & 0.07 & 0.07 \\
click-shades & 0.91 & 1.00 & 0.04 & 0.27 & 0.22 & n/a & n/a & 0.99 & n/a & 0.27 & 0.99 \\
click-shape & 0.88 & 0.95 & 0.11 & 0.11 & 0.64 & n/a & n/a & 0.64 & n/a & 0.64 & 0.64 \\
click-tab-2-easy & 0.99 & 0.99 & 0.61 & n/a & n/a & n/a & n/a & n/a & n/a & n/a & n/a \\
click-tab-2-hard & 0.96 & 0.98 & 0.19 & n/a & n/a & n/a & n/a & n/a & n/a & n/a & n/a \\
click-tab-2-medium & 0.97 & 0.99 & 0.54 & n/a & n/a & n/a & n/a & n/a & n/a & n/a & n/a \\
click-tab-2 & 0.97 & 0.98 & 0.27 & 0.08 & 0.64 & n/a & 1.00 & 0.98 & n/a & 1.00 & 1.00 \\
click-tab & 0.99 & 1.00 & 0.95 & 0.97 & 0.55 & 1.00 & 1.00 & 1.00 & 1.00 & 1.00 & 1.00 \\
click-test-2 & 0.99 & 1.00 & 0.95 & 0.83 & 1.00 & n/a & 1.00 & 1.00 & n/a & 1.00 & 1.00 \\
click-test-transfer & 0.99 & 1.00 & 0.94 & n/a & n/a & n/a & n/a & n/a & n/a & n/a & n/a \\
click-test & 1.00 & 1.00 & 1.00 & 1.00 & 1.00 & n/a & 1.00 & 1.00 & n/a & 1.00 & 1.00 \\
click-widget & 0.83 & 1.00 & 0.56 & 0.34 & 0.93 & n/a & 1.00 & 0.93 & n/a & 1.00 & 1.00 \\
copy-paste-2 & 0.94 & 0.63 & 0.01 & 0.00 & n/a & n/a & n/a & n/a & n/a & 0.00 & 0.00 \\
copy-paste & 0.94 & 0.79 & 0.04 & 0.00 & n/a & n/a & n/a & n/a & n/a & 0.00 & 0.00 \\
count-shape & 0.82 & 0.85 & 0.21 & 0.18 & 0.59 & n/a & n/a & 0.76 & n/a & 0.59 & 0.76 \\
count-sides & 0.98 & 1.00 & 0.74 & 0.30 & n/a & n/a & n/a & n/a & n/a & 0.30 & 0.30 \\
drag-box & 0.99 & 1.00 & 0.61 & 0.31 & n/a & n/a & n/a & n/a & n/a & 0.31 & 0.31 \\
drag-cube & 0.99 & 0.79 & 0.23 & 0.18 & n/a & n/a & n/a & n/a & n/a & 0.18 & 0.18 \\
drag-item & 0.98 & 1.00 & 0.61 & n/a & n/a & n/a & n/a & n/a & n/a & n/a & n/a \\
drag-items-grid & 0.87 & 0.98 & 0.05 & 0.01 & n/a & n/a & n/a & n/a & n/a & 0.01 & 0.01 \\
drag-items & 0.93 & 0.99 & 0.13 & 0.41 & n/a & n/a & n/a & n/a & n/a & 0.41 & 0.41 \\
drag-shapes & 0.96 & 0.99 & 0.26 & 0.92 & n/a & n/a & n/a & n/a & n/a & 0.92 & 0.92 \\
drag-sort-numbers & 0.92 & 0.97 & 0.11 & 0.66 & n/a & n/a & n/a & n/a & n/a & 0.66 & 0.66 \\
email-inbox-delete & 0.99 & 1.00 & 0.22 & n/a & n/a & n/a & 1.00 & n/a & n/a & 1.00 & 1.00 \\
email-inbox-forward-nl-turk & 0.88 & 1.00 & 0.00 & n/a & n/a & n/a & n/a & n/a & n/a & n/a & n/a \\
email-inbox-forward-nl & 0.91 & 1.00 & 0.00 & n/a & n/a & n/a & n/a & n/a & n/a & n/a & n/a \\
email-inbox-forward & 0.96 & 1.00 & 0.01 & n/a & n/a & n/a & n/a & n/a & n/a & n/a & n/a \\
email-inbox-important & 0.99 & 1.00 & 0.30 & n/a & n/a & n/a & n/a & n/a & n/a & n/a & n/a \\
email-inbox-nl-turk & 0.93 & 1.00 & 0.05 & n/a & 0.77 & n/a & n/a & 0.93 & n/a & 0.77 & 0.93 \\
email-inbox-noscroll & 0.96 & 1.00 & 0.13 & n/a & n/a & n/a & n/a & n/a & n/a & n/a & n/a \\
email-inbox-reply & 0.91 & 1.00 & 0.00 & n/a & n/a & n/a & n/a & n/a & n/a & n/a & n/a \\
email-inbox-star-reply & 0.95 & 1.00 & 0.11 & n/a & n/a & n/a & n/a & n/a & n/a & n/a & n/a \\
email-inbox & 0.96 & 1.00 & 0.09 & 0.03 & 0.43 & n/a & 0.54 & 0.99 & n/a & 0.54 & 0.99 \\
enter-date & 0.97 & 1.00 & 0.02 & 0.61 & 0.00 & 1.00 & n/a & 0.96 & 1.00 & 1.00 & 1.00 \\
enter-password & 0.96 & 1.00 & 0.02 & 0.00 & 0.99 & 1.00 & 1.00 & 1.00 & 1.00 & 1.00 & 1.00 \\
enter-text-2 & 0.91 & 0.98 & 0.04 & 0.00 & n/a & n/a & n/a & n/a & n/a & 0.00 & 0.00 \\
enter-text-dynamic & 0.97 & 1.00 & 0.39 & 1.00 & 1.00 & 1.00 & 1.00 & 1.00 & 1.00 & 1.00 & 1.00 \\
enter-text & 0.98 & 1.00 & 0.35 & 0.00 & 1.00 & n/a & 1.00 & 1.00 & n/a & 1.00 & 1.00 \\
enter-time & 0.98 & 0.97 & 0.04 & 0.08 & 0.52 & n/a & n/a & 0.90 & n/a & 0.52 & 0.90 \\
find-midpoint & 0.94 & 0.97 & 0.35 & 0.31 & n/a & n/a & n/a & n/a & n/a & 0.31 & 0.31 \\
find-word & 0.96 & 0.88 & 0.05 & 0.00 & n/a & n/a & n/a & n/a & n/a & 0.00 & 0.00 \\
focus-text-2 & 0.99 & 1.00 & 0.96 & 0.83 & 1.00 & n/a & 1.00 & 1.00 & n/a & 1.00 & 1.00 \\
focus-text & 1.00 & 1.00 & 0.99 & 0.95 & 1.00 & n/a & 1.00 & 1.00 & n/a & 1.00 & 1.00 \\
grid-coordinate & 0.87 & 1.00 & 0.66 & 0.26 & 1.00 & n/a & n/a & 1.00 & n/a & 1.00 & 1.00 \\
guess-number & 0.99 & 1.00 & 0.21 & 0.20 & 0.00 & n/a & n/a & 0.00 & n/a & 0.20 & 0.20 \\
highlight-text-2 & 0.97 & 1.00 & 0.40 & 0.13 & n/a & n/a & n/a & n/a & n/a & 0.13 & 0.13 \\
highlight-text & 0.97 & 1.00 & 0.51 & 0.90 & n/a & n/a & n/a & n/a & n/a & 0.90 & 0.90 \\
identify-shape & 0.98 & 1.00 & 0.68 & 0.36 & 0.90 & n/a & n/a & 1.00 & n/a & 0.90 & 1.00 \\
login-user-popup & 0.94 & 1.00 & 0.02 & n/a & n/a & n/a & n/a & n/a & n/a & n/a & n/a \\
login-user & 0.96 & 1.00 & 0.00 & 0.00 & 0.99 & 1.00 & 1.00 & 1.00 & 1.00 & 1.00 & 1.00 \\
moving-items & 0.18 & 0.88 & 0.13 & 0.78 & n/a & n/a & n/a & n/a & n/a & 0.78 & 0.78 \\
multi-layouts & 0.95 & 1.00 & 0.00 & n/a & 0.99 & n/a & n/a & 1.00 & n/a & 0.99 & 1.00 \\
multi-orderings & 0.96 & 1.00 & 0.00 & n/a & 0.05 & n/a & n/a & 1.00 & n/a & 0.05 & 1.00 \\
navigate-tree & 0.98 & 0.99 & 0.32 & 0.20 & 0.99 & 1.00 & 1.00 & 0.99 & 1.00 & 1.00 & 1.00 \\
number-checkboxes & 0.96 & 0.99 & 0.00 & 0.16 & n/a & n/a & n/a & n/a & n/a & 0.16 & 0.16 \\
read-table-2 & 0.95 & 0.94 & 0.00 & 0.00 & n/a & n/a & n/a & n/a & n/a & 0.00 & 0.00 \\
read-table & 0.97 & 0.97 & 0.01 & 0.00 & n/a & n/a & n/a & n/a & n/a & 0.00 & 0.00 \\
resize-textarea & 0.94 & 1.00 & 0.27 & 0.11 & n/a & n/a & n/a & n/a & n/a & 0.11 & 0.11 \\
right-angle & 0.87 & 0.98 & 0.26 & 0.38 & n/a & n/a & n/a & n/a & n/a & 0.38 & 0.38 \\
scroll-text-2 & 0.97 & 1.00 & 0.88 & 0.96 & n/a & n/a & n/a & n/a & n/a & 0.96 & 0.96 \\
scroll-text & 0.97 & 0.96 & 0.04 & 0.00 & n/a & n/a & n/a & n/a & n/a & 0.00 & 0.00 \\
search-engine & 0.97 & 1.00 & 0.15 & 0.00 & 0.26 & n/a & 1.00 & 0.99 & n/a & 1.00 & 1.00 \\
simon-says & 0.62 & -0.00 & 0.02 & 0.28 & n/a & n/a & n/a & n/a & n/a & 0.28 & 0.28 \\
simple-algebra & 0.86 & 0.75 & 0.03 & 0.04 & n/a & n/a & n/a & n/a & n/a & 0.04 & 0.04 \\
simple-arithmetic & 0.96 & 0.86 & 0.38 & 0.07 & n/a & n/a & n/a & n/a & n/a & 0.07 & 0.07 \\
social-media-all & 0.89 & 0.75 & 0.00 & n/a & 0.01 & n/a & n/a & 0.01 & 1.00 & 0.01 & 1.00 \\
social-media-some & 0.91 & 0.85 & 0.01 & n/a & 0.01 & n/a & n/a & 0.42 & n/a & 0.01 & 0.42 \\
social-media & 0.96 & 0.90 & 0.03 & 0.23 & 0.39 & n/a & 1.00 & 1.00 & n/a & 1.00 & 1.00 \\
terminal & 0.88 & -0.01 & 0.00 & 0.00 & n/a & n/a & n/a & n/a & n/a & 0.00 & 0.00 \\
text-editor & 0.88 & 0.98 & 0.11 & 0.01 & n/a & n/a & n/a & n/a & n/a & 0.01 & 0.01 \\
text-transform & 0.86 & 0.60 & 0.19 & 0.00 & n/a & n/a & n/a & n/a & n/a & 0.00 & 0.00 \\
tic-tac-toe & 0.71 & 0.83 & 0.32 & 0.34 & 0.37 & n/a & n/a & 0.47 & n/a & 0.37 & 0.47 \\
unicode-test & 0.99 & 1.00 & 0.86 & n/a & n/a & n/a & n/a & n/a & n/a & n/a & n/a \\
use-autocomplete & 0.98 & 1.00 & 0.07 & 0.00 & 0.78 & n/a & n/a & 0.98 & n/a & 0.78 & 0.98 \\
use-colorwheel-2 & 0.94 & 0.95 & 0.38 & 1.00 & n/a & n/a & n/a & n/a & n/a & 1.00 & 1.00 \\
use-colorwheel & 0.90 & 0.98 & 0.68 & 1.00 & n/a & n/a & n/a & n/a & n/a & 1.00 & 1.00 \\
use-slider-2 & 0.97 & 0.95 & 0.03 & 0.15 & n/a & n/a & n/a & n/a & n/a & 0.15 & 0.15 \\
use-slider & 0.98 & 0.91 & 0.18 & 0.51 & n/a & n/a & n/a & n/a & n/a & 0.51 & 0.51 \\
use-spinner & 0.98 & 1.00 & 0.47 & 0.17 & 0.04 & n/a & n/a & 0.04 & n/a & 0.17 & 0.17 \\
visual-addition & 0.97 & 0.99 & 0.36 & 0.01 & n/a & n/a & n/a & n/a & n/a & 0.01 & 0.01 \\
\hline
\end{longtable}
\end{tiny}

\begin{figure*}
    \centering
    \includegraphics[width=0.86\textwidth]{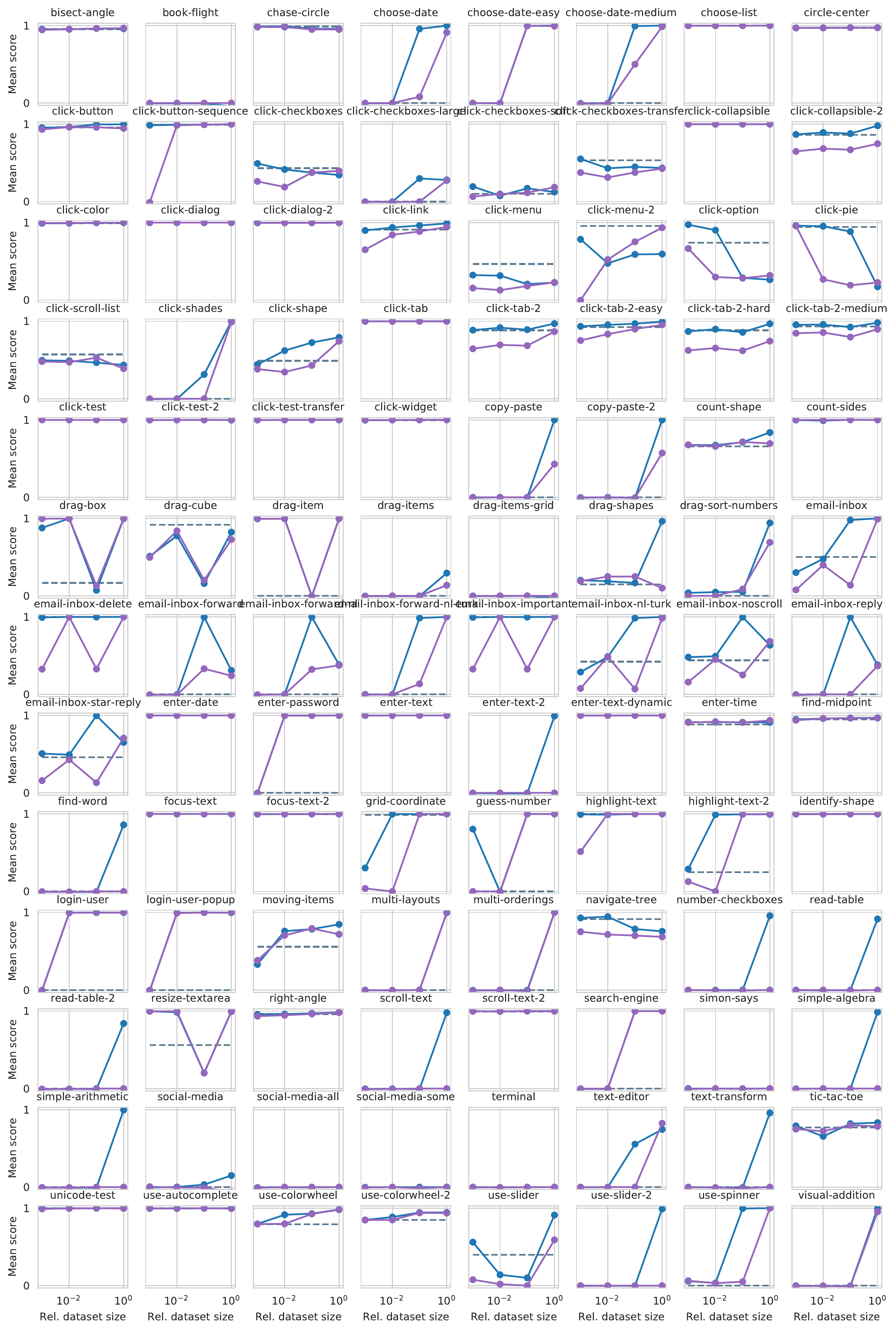}
    \caption{Results from Fig.~\ref{fig:dataset_scaling} broken out per task. Note that the individual task results have significant variance as there is only one seed per experiment. Blue shows performance of BC\&RL, purple BC only, and the dashed line gives RL only performance.}
    \label{fig:per_task_dataset_ablation}
\end{figure*}

\begin{figure}
    \centering
    \includegraphics[width=0.7\textwidth]{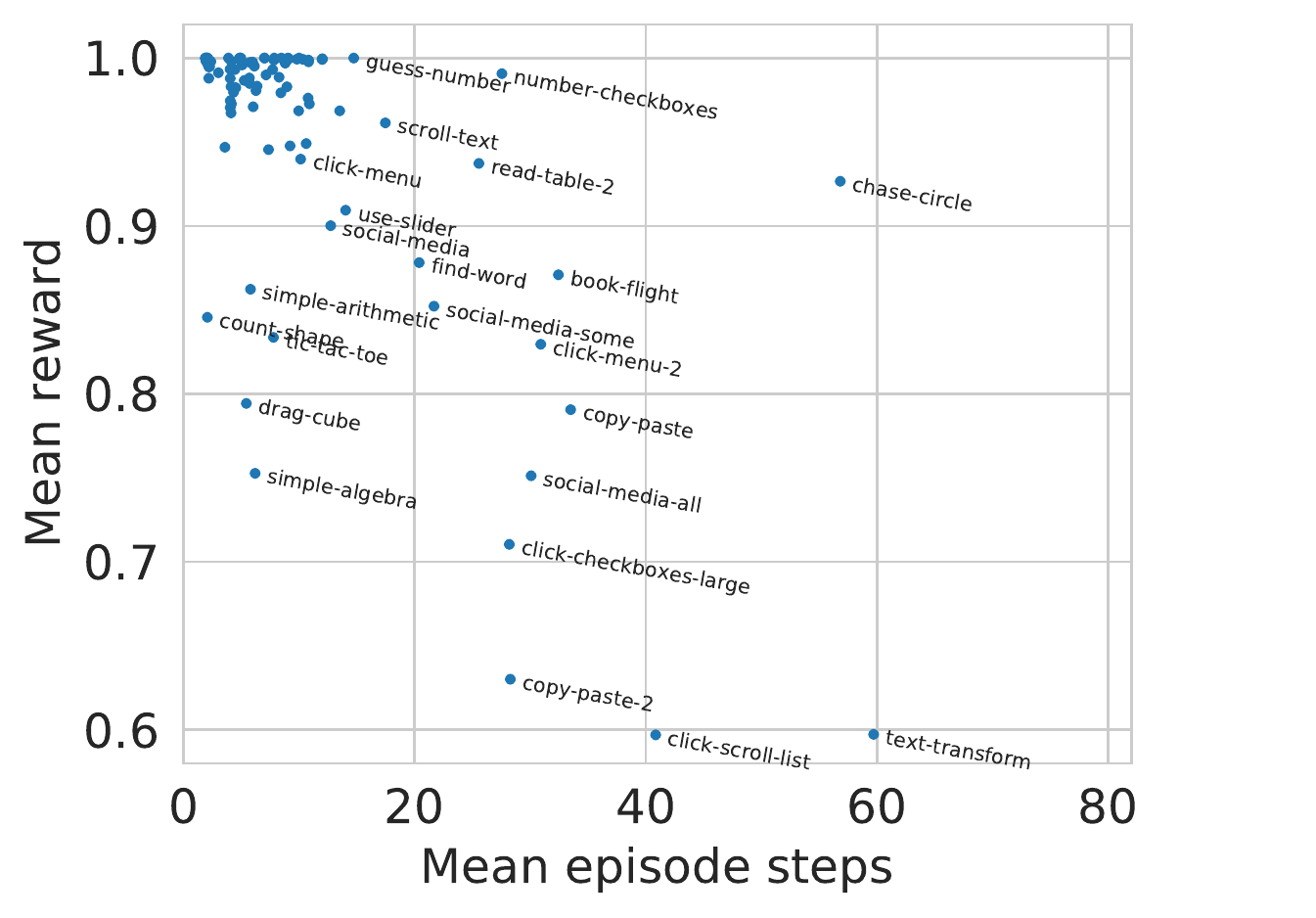}
    \caption{Scatter plot showing the mean number of environment steps per episode versus mean reward for different MiniWob tasks.}
    \label{fig:per_task_dataset_ablation}
\end{figure}

\end{document}